\documentclass{article}

\usepackage[english]{babel}

\usepackage[letterpaper,top=2cm,bottom=2cm,left=3cm,right=3cm,marginparwidth=1.75cm]{geometry}

\usepackage{amsmath}
\usepackage{graphicx}
\usepackage[colorlinks=true, allcolors=blue]{hyperref}
\usepackage{hyperref}
\usepackage{url}
\usepackage{array}
\usepackage{algorithm}
\usepackage{algpseudocode}
\graphicspath{ {./Images/} }
\usepackage[table]{xcolor}
\usepackage[aboveskip=1pt,labelfont=bf,labelsep=period,justification=raggedright,singlelinecheck=off]{caption}
\usepackage{lastpage,fancyhdr,graphicx}
\usepackage{epstopdf}
\usepackage{amsmath,amssymb}
\usepackage{textcomp,marvosym}
\usepackage{changepage}
\title{Unsupervised Anomaly Appraisal of Cleft Faces Using a StyleGAN2-based Model Adaptation Technique}

\begin{document}
\maketitle
\begin{flushleft}
\author{Abdullah Hayajneh\textsuperscript{1},
Mohammad Shaqfeh\textsuperscript{2},
Erchin Serpedin\textsuperscript{1},
Mitchell A. Stotland\textsuperscript{3}
\\
\textbf{1}  Electrical and Computer Engineering Department, Texas A\&M University, College Station, TX, USA
\\
\textbf{2} Electrical and Computer Engineering Program, Texas A\&M University, Doha, Qatar
\\
\textbf{3} Division of Plastic, Craniofacial and Hand Surgery, Sidra Medicine, and Weill Cornell Medical College, Doha, Qatar
}
\end{flushleft}
\begin{abstract}
This paper presents a novel machine learning framework to consistently detect, localize and rate congenital cleft lip anomalies in human faces. The goal is to provide a universal, objective measure of facial differences and reconstructive surgical outcomes that matches human judgments. The proposed method employs the StyleGAN2 generative adversarial network with model adaptation to produce normalized transformations of cleft-affected faces in order to allow for subsequent measurement of deformity using a pixel-wise subtraction approach (as illustrated in Figure \ref{img:result}). The complete pipeline of the proposed framework consists of the following steps: image preprocessing, face normalization, color transformation, morphological erosion, heat-map generation and abnormality scoring. Heatmaps that finely discern anatomic anomalies are proposed by exploiting the features of the considered framework. The proposed framework is validated through computer simulations and surveys containing human ratings. The anomaly scores yielded by the proposed computer model correlate closely with the human ratings of facial differences, leading to  0.942 Pearson's r score.
\end{abstract}

\begin{figure}[h]
\begin{center}
\includegraphics[width=12cm]{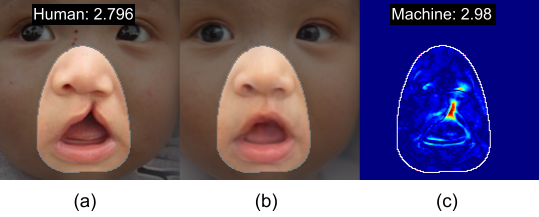}
\end{center}
\caption{\label{img:result}Anomaly scores obtained using the proposed framework agrees with the human scores:  (a) Original anomalous  face  \cite{1png} with its human survey score, (b) Normalized version obtained using the StyleGAN2-based model adaptation technique, and (c) Heatmap generated by the proposed framework with its calculated score.}
\end{figure}


\section*{Introduction}

The detection of anomalies is a complementary component of pattern recognition, and an integral function of machine learning models. Our research is concentrated on designing an automated system that can both objectively detect anomalous structures within a human face as well as rate the extent of anomaly in a manner analogous to human judgement. In the current protocol, we have focused on cleft lip deformity which is a common birth defect occurring in approximately 1/1600 live births in the United States \cite{mai2019national}. Patients affected by cleft lip commonly undergo multiple surgical interventions over the course of their childhood in an effort to diminish evidence of abnormality. Decision-making regarding timing, extent, and outcome of surgical reconstruction is primarily based on subjective assessment on the part of the patient, parents, and provider. The management of congenital and acquired forms of facial difference would benefit considerably from a universally accessible and standardized form of measurement. Creating an automated scale with which to gauge the human face would help inform patient discussion, refine surgical planning and outcome measurement, and better characterize the clinical need and benefit of surgery for third-party payers.

The task of the designed system is to detect the existence of structural outliers within a face and then to provide an indicator of magnitude. Such a system of measurement would ideally demonstrate (i) order-preservation: reliably ranking anomaly severity within sets of images with a hierarchy corresponding to human appraisal, (ii) indifference to extraneous variation: tolerance to variations in age, gender, race, pose, lighting, etc., and (iii) comprehensibility: revealing an interpretable process of distinguishing subtle zones of facial aberration, e.g., through the provision of corresponding heat detection maps.

Designing anomaly detectors is in general challenged by the unavailability or scarcity of anomalous data sets (i.e., data sets that refer to rare or abnormal conditions). Therefore, building anomaly detectors that can be trained using only normal data profiles represents an attractive solution to overcome this challenge. Furthermore, in order to build anomaly detectors with high sensitivity and specificity, the detectors can be equipped with deep architectures to efficiently model the complex patterns and correlations embedded into the data sets associated with normality regimes in order to effectively distinguish abnormal conditions as outliers.

\begin{figure}[h]
\begin{center}
\fbox{\includegraphics[width=8cm]{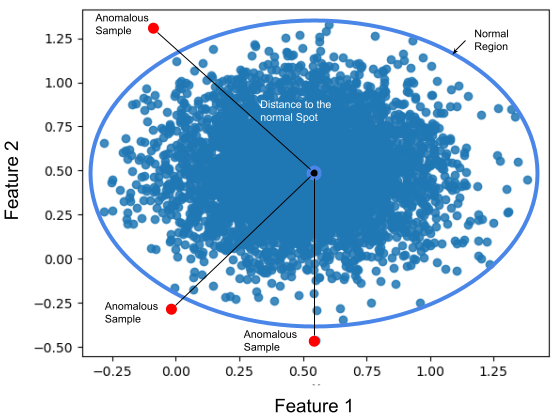} }
\end{center}
\caption{\label{img:anomaly_detection_approaches} Anomaly detection methods bound the space occupied by normal spots and calculate the distance between anomalous samples that are outside these boundaries and the center to the spots.}
\end{figure}

Methods for anomaly detection have been proposed for a wide range of  medical applications, such as identification of abnormalities in airways   \cite{jiao2013detection}, chest radiographs  \cite{nakao2021unsupervised}, retinal tomographic images  \cite{schlegl2017unsupervised}, and mammography and thyroid data sets \cite{kiat2018doping}. Most of the anomaly detection methods depend on explicitly modeling the distribution of the normal data after being transformed into a feature space, then detecting the anomalous sample according to how far it is located from the spots occupied by the normal samples in the feature space as illustrated in Figure \ref{img:anomaly_detection_approaches}.  
The work  \cite{erfani2016high} used one-class Support Vector Machines (SVMs) trained on one category of samples that represent features offered by a trained deep belief network (DBN) model. This model transforms the original one-class samples into another feature space where the transformed normal samples can be better discriminated from  other transformed samples including  anomalous samples. This approach was applied on different low dimensionality datasets $(< 600)$. However, it was not tested on samples of higher dimensions (e.g., 1024 $\times$ 1024 images). A different approach employs  a Random Forest (RF) classifier \cite{ho1995random} trained on well-selected features to distinguish the elderly people whose eyes are affected by AMD (Age-related Macular Degeneration)  \cite{venhuizen2015automated}. Reference  \cite{seebock2016identifying} uses a variational autoencoder (VAE) \cite{kingma2013auto} with a one-class SVM model to identify anomalous regions in a retinal imaging dataset. 

The recent advances in the field of  Generative Adversarial Networks  (GANs)\cite{goodfellow2020generative} offer  new perspectives to design powerful anomaly detection and localization approaches. AnoGAN framework in \cite{nakao2021unsupervised} generates an anomaly score to rate the health status of the retina and to localize the anomalous locations using tomography images. The AnoGAN model was trained from scratch using 1M patches to generate new and unique samples. The closest generated sample to the one under analysis is obtained by finding an optimized latent vector corresponding to the required generated sample. The loss function employed for this optimization uses both the generator and the discriminator of the GAN. Then, a heatmap is generated by defining the residual image as the pixel-wise difference between the generated image and the real one. The scoring system was built based on the loss function used during the latent mapping operation. 

Another utilization of GANs was carried out in \cite{boyaci2020personalized}. This approach uses the StyleGAN \cite{karras2019style} face generator to normalize and score anomalous faces using a similar approach to AnoGAN. Regarding the loss function used for the optimization, it employs a combination of two measures: the “similarity” and “averageness”. The first measure tries to preserve the distinctive features of the generated face, while the second measure tries to keep the face as close to the “average face” as possible. For the scoring, the authors trained a separate network to generate a score for each anomalous face using different extracted features of it. 
One limitation of both previously mentioned works is that the generated samples do not show a perfect normalization of the given real one, as some  details of the sample were not represented in the pretrained model. Furthermore, the generated scores do not align well with the human scores and the time required to score one image is long. 

The contributions of this paper are the following:
\begin{itemize}
  \item a novel StyleGAN2-based model adaptation algorithm to fine-tune normalized cleft faces by incorporating additional identity-preserving facial features  relative to the standard StyleGAN2 projection based algorithm. 
  \item a heat map generation framework that employs a residual image,  yielded by an image processing pipeline, and it is used to compare an input raw image with a normalized counterpart.
  \item a heat map-based face scoring system to evaluate and sort samples according to their closeness to normality.

\end{itemize}

\section{Materials and Methods}
\begin{figure}[h]
\begin{center}
\fbox{\includegraphics[width=13.5cm]{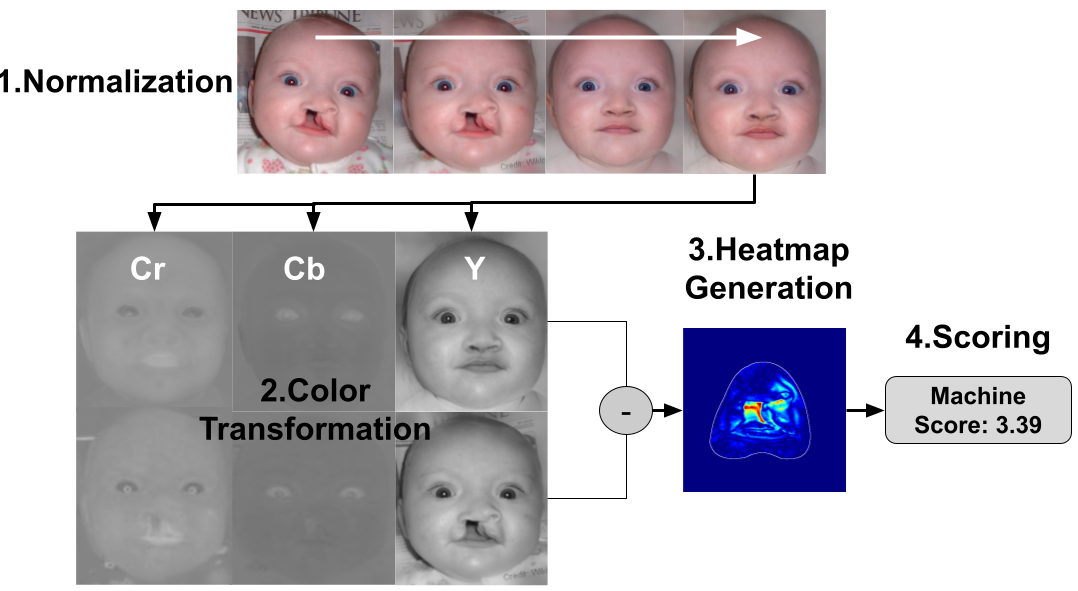} }   
\end{center}
\caption{Overall scoring system for a given face \cite{2png}. It assumes multiple processing steps: image preprocessing, face normalization, color transformation, morphological erosion, heat map calculation and anomaly scoring.  \label{fig3}}
\end{figure}
The proposed method aims to measure the difference between the abnormal face image and a normalized version of it to create an overall score that describes the severity of the anomaly.  The key idea of our method is to utilize the capabilities offered by the StyleGAN2 facial generator \cite{karras2020analyzing} to normalize the face and the pixel wise subtraction technique without being affected by the image changes caused by noise. The noise may come from two different sources. The first source is from the image acquisition device. The other source is from the StyleGAN2 generator that adds a specific level of structural artifacts when producing a face \cite{tan2021systematic}.

The proposed methodology depicted in Figure \ref{fig3} consists of the following steps: Image Preprocessing, Image Normalization, Color Transformation, Morphological Erosion, Heat Map Generation via pixel wise subtraction, and finally Anomaly Scoring. Next,  we will discuss each step separately. 

\subsection{Image Preprocessing}
The image preprocessing steps are very important for the success of the proposed method. All the machine learning models used in the proposed model are designed to accept facial images with specific scale and orientation. 

In the first preprocessing step, the face inside the image is detected and localized using a well trained face detection model (e.g., Haar classifier \cite{viola2001rapid} or Local Binary Patterns (LBP) \cite{huang2011local}). Then we check if the area ratio between the face and the background is consistent with the CelebaHQ \cite{karras2019style} dataset setup or not. In general, the standard analysis of StyleGAN2 pretrained model requires approximately 40\% of the input image to be background and 60\% face. If the background is very large compared to the area of the face, then majority of it will later be removed. In the other case, if the background in the image is very small, it will be enlarged by blurring the background and place 8 horizontally/vertically flipped replicates of the image around it. This helps to match the shape of the input image with the StyleGAN2 generator input shape (1024 $\times$ 1024 pixels). 

To create a blurred background of the image around the face, a mask image is generated  by separating the foreground from the background pixels of the original face image using the previously detected face. Afterwards, a Gauss smoothing filter \cite{davies2012computer} is applied to the mask to smooth the transition between the foreground and the background areas. Then the range of values of the mask image are converted from 0-255 to 0-1 and multiplied with the original face image.

\begin{figure}[h]
\begin{center}
\fbox{\includegraphics[width=12cm]{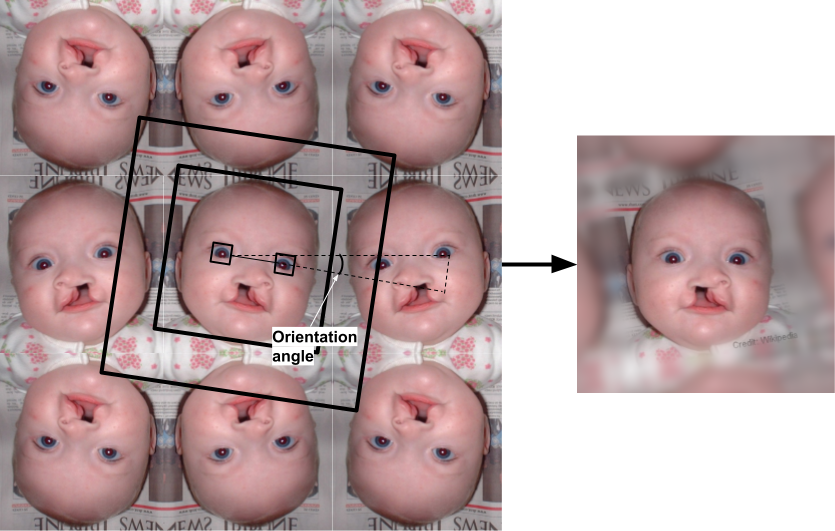} }
\end{center}
\caption{\label{img:preprocessing}Applying different preprocessing steps to generate consistent images with the StyleGAN2 pretrained model. This includes adjusting the background, scale, and orientation.}
\end{figure}

The next preprocessing step is to detect and correct the orientation of the face inside the image. This is conducted by detecting the eyes in the face by using 68 landmarks of the face. The goal is to secure horizontal alignment of the eyes inside the image. Next, the distance between the eyes is measured and the whole image is scaled up or down so that the distance between the centers of the eyes is equal to 100 pixels, a condition  which ensures consistency with the StyleGan2 pretrained model. Finally, 1024 $\times$ 1024 pixels are cropped around the face location. Figure 
\ref{img:preprocessing} illustrates the overall preprocessing steps.  

\subsection{Image Normalization}
\begin{figure}[h]
\begin{center}
\fbox{\includegraphics[width=13cm]{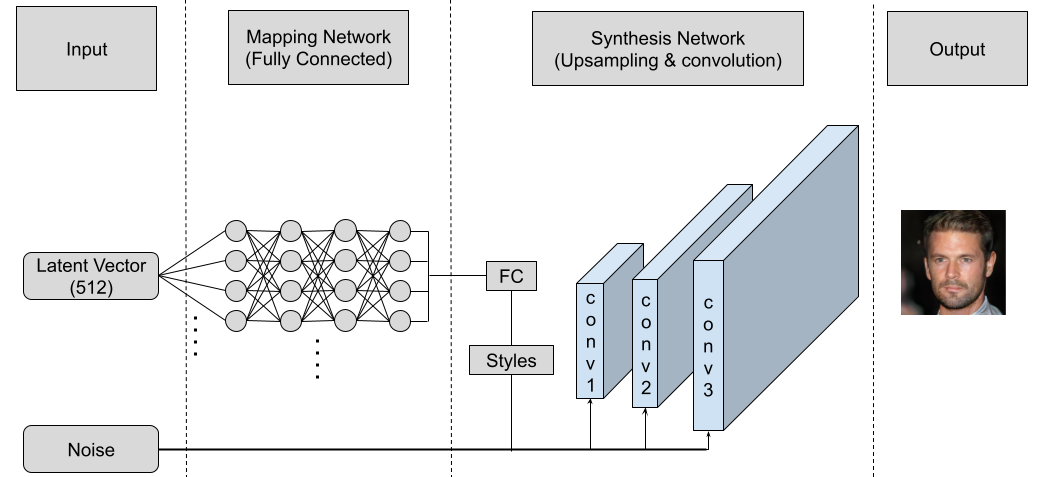} }
\end{center}
\caption{\label{img:StyleGAN} General architecture of the StyleGAN2 generator. It consists of mapping and synthesis networks. The first network facilitates disentangled representation of the object features. The synthesis network produces diverse human faces by using the generated styles and noise maps with different sizes. At the end, this face generator can real-looking faces like the one shown in the right side of the figure \cite{styleGAN2png}  }
\end{figure}

After obtaining a well-aligned face image through the preprocessing step, a normalized version of the face is created using the StyleGAN2 face generator which can produce unique high quality faces. The structure of the StyleGAN2 face generator is depicted in Figure \ref{img:StyleGAN}. The input to the generator is a 512 dimensional latent vector which encodes different features of the face. It transforms the latent vector into a 1024 $\times$ 1024 dimensional face image by passing the latent vector through the mapping and synthesis networks. The aim of the mapping network is to obtain better features representation through an \emph{intermediate} latent space $\mathcal{W}$. Moving in one direction in the latent space should generate a consistent facial appearance change in the corresponding image (e.g., skin color). During the face generation in the synthesis network, styles produced from the latent vector are used along with the  convolutional layers to control the styles of the generated faces including aspects such as skin color, eyes size, mouth width, etc. In addition, random noise maps are employed to allow for more stochastic variations in the image domain.  

\begin{figure}[h]
\begin{center}
\fbox{\includegraphics[width=13cm]{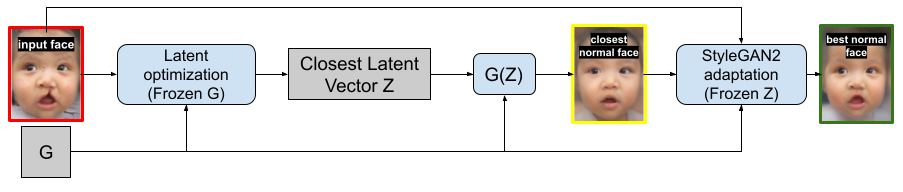} }
\end{center}
\caption{\label{img:Normalization_pipeline} Proposed normalization algorithm. It searches for the closest latent vector closely matching the input face. Then the latent vector is frozen and the StyleGAN2 model is then optimized to reconstruct more facial details without anomalies. G refers to the StyleGAN2 generator network.}
\end{figure}

All of the above mentioned features of StyleGAN2 contribute to developing a good face normalization algorithm with no anomalous parts produced. The overall proposed normalization algorithm is illustrated in Figure \ref{img:Normalization_pipeline}. It generally consist of two successive operations: Latent Optimization and StyleGAN2 Adaptation.

An important question that pertains to  a given face with cleft anomaly is:  how can StyleGAN2 be utilized to find a matching normal face without the cleft mouth anomaly that preserves the identity of face? This will be achieved by the \emph{face inversion} operation which finds the vector in the latent representation of a pretrained StyleGAN2 face generator model that transforms into a guess of the normal face image that most closely resembles the given abnormal face under evaluation. In this way, the method will confirm that all the generated details are normal because the StyleGAN2 model was trained on normal samples only. The reason of naming this step as face inversion is that the process finds a latent vector given a face instead of producing a face given a latent vector. 

Different approaches were proposed in the literature to find the closest latent representation of a face in StyleGAN2. These approaches can be divided into two main categories. The first category is represented by the optimization based methods where a latent code is directly optimized for a fixed sample \cite{karras2020analyzing},\cite{abdal2019image2stylegan},\cite{abdal2020image2stylegan++}, \cite{lipton2017precise}, while the second category comprises the encoder based methods where a separate encoder network is built and trained to predict the latent code for the sample \cite{luo2017learning},\cite{guan2020collaborative}, \cite{perarnau2016invertible}. The encoder based methods present the advantage of producing the corresponding latent vector in a single forward iteration through the encoder. This helps in significantly reducing the process overhead compared to the optimization based methods. On the other hand, the optimization based methods can produce latent vectors associated with better quality images and closer to the input image. 

Additionally, a hybrid method lying between the two above mentioned categories was proposed \cite{zhu2020domain}. First, the hybrid method  finds an estimate of the latent code using a well trained encoder to shorten the latent optimization overhead. Then it employs an iterative optimization algorithm   to refine the latent code and to better represent the semantics of the image in the latent space.
\subsubsection{StyleGAN2 Latent Optimization - Face Inversion}
Having mentioned different face inversion methods, anyone of them can achieve the required aim of this step. For the sake of consistency in the overall pipeline, this step uses the standard face inversion method proposed by the StyleGAN2 original paper and described in Algorithm \ref{alg:inversion}. The chosen inversion method resides in the category of the optimization based methods. The key advantage of this framework is that it additionally optimizes the 18 noise maps $\textbf{n}_i$, $i \in {0,1, ...,17},$ present in the StyleGAN2 architecture to help in finding the best latent vector. In general, this method iteratively calculates and minimizes a similarity loss between the input face image and a randomly generated face by adjusting a random latent vector using the back-propagation method \cite{rumelhart1986learning}. The utilized loss function is based on the Learned Perceptual Image Patch Similarity (LPIPS) \cite{zhang2018unreasonable} metric $\mathcal{D}_{LPIPS}$. LPIPS is the state-of-the-art image comparison metric which calculates the averaged difference between deep features extracted from two images using feature extractor networks such as pretrained VGG \cite{simonyan2014very}, Alexnet \cite{krizhevsky2017imagenet} or SqueezeNet \cite{iandola2016squeezenet}. The LPIPS loss function takes  the following mathematical expression: 

\begin{equation}
\begin{array}{l}

\label{eqn:eqlabel}
\mathcal{D}_{LPIPS}(x,y)=\sum_l\frac{1}{W_lH_l}\sum_{h,w}^{W_l,H_l}||w_l\bigodot(\hat{x}^l_{hw}-\hat{y}^l_{hw})||_2^2  \;  , 
\end{array}
\end{equation}
where $l$ stands for  the layer index, $w_l$, $W_l$ and $H_l$  denote  the the learned weights, width and height of the layer $l$, respectively, and $\bigodot$ represents the Hadamard product (vector element-wise  multiplication).

The following summarizes the original StyleGAN2 inversion algorithm.  It starts by generating 10,000 random latent vectors $Z$, transforming them using the mapping network to produce 10,000 intermediate latent vectors $W$ and averaging them to get an estimate of the mean intermediate latent vector $\mu$. The algorithm optimizes $\mu$ to get an initial guess of the latent vector corresponding to the closest generated face to the input image. The optimization is done by calculating the similarity loss between the original image and the guess using the LPIPS distance measure $\mathcal{D}_{LPIPS}$. To facilitate the latent vector search,  the inversion algorithm additionally optimizes  18 randomly generated noise maps with different resolutions $\textbf{n}_i \in \mathcal{R}^{r_i\times r_i}$, $r_i \in \{1024,512,...,8\}$, for face synthesis operation. As a consequence of optimizing the noise maps, some facial details may sneak into the noise maps. This breaks the assumption that the noise maps have to be random. Therefore, a regularization term $\mathcal{L}_{i,j} $ scaled by $\alpha$ is added to the overall loss, which depends on the original noise maps as well as  scaled versions of noise maps  $\textbf{n}_{i,j}, j>0$.  These scaled versions are not part of the optimization. The $\mathcal{L}_{i,j} $ regularization term measures the amount of randomness present in the noise maps and makes sure that the noise maps are still random.

\begin{algorithm}
\caption{\label{alg:inversion}Face Inversion, Input: Face image $x$ and StyleGAN2 generator model $G$. Output: Latent vector $w$ and set of noise maps $n_i , i=18, \alpha=10^5$}\label{alg:inversion}
\begin{algorithmic}
\State $Z=\{Z_1, ...Z_{10000}\} \gets U\{0,1,..,10000\}$
\State $W=\{W_1,..,W_{10000}\} \gets G(Z)$
\State $\mu \gets \frac{\sum_i{(W_i)}}{N}$
\State $\{\textbf{n}_1, \textbf{n}_2, ...\textbf{n}_i\} \gets \{\mathcal{R}^{r_1\times r_1},\mathcal{R}^{r_2\times r_2},..,\mathcal{R}^{r_i\times r_i}\}$
\While{not converge}
\State $\mathcal{L}_{Image}=\mathcal{D}_{LPIPS}(x,G(\mu,\textbf{n}_i,..))$
\State $n_{i,j} \gets Downsample(n_i,0.5), 0\le j<i,j \in \mathbb{N}$
\State $\mathcal{L}_{i,j} \gets (\frac{1}{r_{i,j}^2} 
\sum_{x,y}n_{i,j}(x,y)  n_{i,j}(x-1,y))^2+ (\frac{1}{r_{i,j}^2} 
\sum_{x,y}n_{i,j}(x,y)  n_{i,j}(x,y-1))^2$
\State $\mathcal{L}(x,G(W,\textbf{n}_i,..))=\mathcal{L}_{Image}+\alpha \sum_{i,j} \mathcal{L}_{i,j}$
\State $\nabla w \gets \frac{d \mathcal{L}} {d w}$
\While{$i$ in $(0,1,..,18)$}
\State $\nabla \textbf{n}_i \gets \frac{d \mathcal{L}} {d \textbf{n}_i}$
\State $\textbf{n}_i= \textbf{n}_i+\nabla \textbf{n}_i $
\EndWhile

\State $w= w+\nabla w $
\EndWhile
\State \Return $w,\textbf{n}_1, \textbf{n}_2, ...\textbf{n}_i$
\end{algorithmic}
\end{algorithm}

\subsubsection{ StyleGAN2 Pretrained Model Adaptation}
After conducting the estimation of the normalized face during the inversion step, we observed that the obtained estimate still lacks some identity preserving details of the abnormal face.  This issue may result in the failure of the abnormality evaluation step. The lack of the reconstruction details is due to either the generator that may not contain representations for all the real faces under evaluation or to the searching method that may fail to explore the entire latent space. This prompts for additional training of the StyleGAN2 pretrained model during this step. The additional training of the StyleGAN2 model is called model adaptation, a step in which the  generator internal weights are updated and forced to represent the remaining details of the input face and to improve the face image. A  custom adaptation method is described in Algorithm \ref{alg:adaptation}. The core of this algorithm is to incrementally adjust the StyleGAN2 weights and freeze the initial estimation of the latent vector obtained from the previous step. The initial guess of the latent vector  corresponds to the best normalized face. This will add more identity-preserving details in the generated face except those related to the anomaly. This step continues for a proper number of training iterations and stops when most of the details are represented in the generated face.  
\begin{algorithm}
\caption{Pretrained model adaptation. Input: Face image $x$, its corresponding closest latent vector $z$ and a StyleGAN2 generator $G$. Output: Adapted generator $G’$}\label{alg:adaptation}
\begin{algorithmic}
\State $G' \gets G$
\While{not converge}
\State $x' \gets G'(z)$
\State $\mathcal{L}(x,x')=\mathcal{L}_{LPIPS}(x,x')+\mathcal{L}_{L2}(x,x')$

\State $\nabla g \gets \frac{d \mathcal{L}} {d z}$
\State $G'= G'+\nabla g $
\EndWhile
\State \Return $G';x'$
\end{algorithmic}
\end{algorithm}
To consistently represent more details of the face, a carefully designed loss function is used during the adaptation step as shown in the following equation:
\begin{equation}
    \mathcal{L}(x_{org},x_{norm})=\mathcal{D}_{LPIPS}(x_{org},x_{norm})+\mathcal{D}_{L2}(x_{org},x_{norm})  .
\end{equation}

\begin{figure}[h]
\begin{center}
\fbox{\includegraphics[width=13.5cm]{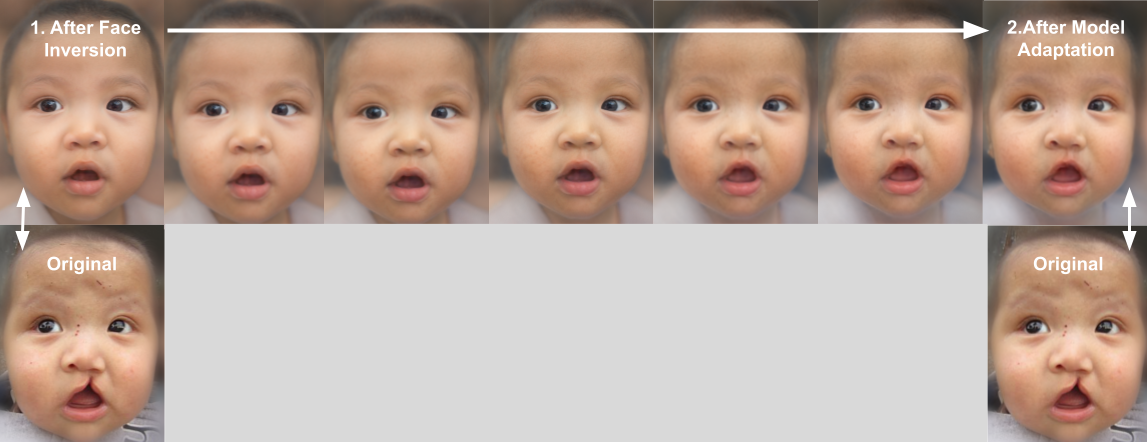} }
\end{center}
\caption{\label{img:overall_normalization}Face transformation when applying the face inversion and the model adaptation algorithms sequentially.}
\end{figure}

The above loss function tries to compromise between reconstructing semantic details of the face captured by $\mathcal{D}_{LPIPS}$ distance measure, and the details of image pixels present in both the original and generated faces indicated by $\mathcal{D}_{L2}$. 
This loss function is used to incrementally update the weights of StyleGAN2 to get the new adapted generator $G'$ which contains the closest possible face $x'$ to the original one $x$. An example of the complete image normalization pipeline is described in Figure \ref{img:overall_normalization}.

\begin{figure}[h]
\begin{center}
\fbox{\includegraphics[width=12cm]{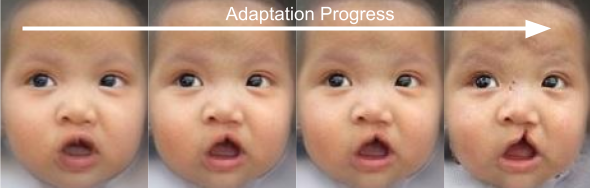} }
\end{center}
\caption{\label{img:adaptation}If we keep adapting the StyleGAN2 generator for more than 50 iterations, the model starts reconstructing the abnormal details of the original face. Here the cleft anomaly gradually appears in the generated face}
\end{figure}

As mentioned before, selecting the number of adaptation iterations depends on the amount of missing details we want  to be present. If the input face is very close to normal, then less adaptation iterations are required, as the majority of the face details will be present in the first inversion step. A proper number of iterations to obtain most of the normal missing details of the generated face was found to be 50 adaptation iterations. If the adaptation proceeds with more iterations than this number, abnormal parts of the input face will be gradually  reconstructed which harms the normalization result, as shown in Figure \ref{img:adaptation}. Nevertheless, this observation can be utilized in other studies such as simulation of the recovery state of a cleft affected patient. 
Also,  extra iterations  can be applied to  fully reconstruct the anomalous parts and the number of iterations depends on the degree of severity of the anomaly.  The strong correlation between the number of iterations and degree of anomaly severity can be employed  to generate an alternative severity index for facial deformities, a research direction that we plan addressing in a future companion  paper.

 Several works have been proposed to adapt new datasets into a pretrained StyleGAN2 model \cite{wang2018transferring}, \cite{mo2020freeze}, \cite{noguchi2019image}, \cite{karras2020training}. Most of them are designed to change the domain of the generated objects. Our approach benefits from these so that a small update can be applied to the generator to produce the required details of the original face. This small update does not alter the remaining latent space and its corresponding generated images. Also, we do not utilize the edited generator for any other task except to refine the target normalized face.

\subsection{Color Transformation}
Returning to our goal, we aim to generate an anomaly difference map between the abnormal face and the normalized one. This map should highlight the abnormal location in the face but not all the other normal differences like small color changes due to lighting. In the majority of cases, the anomaly part appears as a geometrical change in the image more than color change. The geometrical information of the image can be described more precisely using the intensity information than the color information. This requires a representation of the image using a color system that separates the color information from the intensity.

The Red-Green-Blue (RGB) color model describes a pixel information by the amount of each of the three colors included in it. The range of each color value is not linearly related to the intensity. Also, the RGB color system presents a lot of visual information redundancy in its three channels with no separation between color and intensity.

YCbCr is another color representation model that presents the advantage of separating the intensity values from chromaticity. In this color model, Y represents the intensity, Cb denotes the blue difference and Cr stands for the red difference \cite{koschan2008digital}. Conversion from RGB to YCbCr color model (after gamma correction) is obtained using these transformations: 

\begin{equation}
\begin{array}{l}

\label{eqn:eqlabel}
Y'=16+(65.481\cdot R'+128.553\cdot G'+24.966\cdot B')\\

C_B=128+(-37.797\cdot R'+74.203\cdot G'+112.0\cdot B')
\\
C_R=128+(112.0\cdot R'+93.786\cdot G'+18.214\cdot B')
\end{array}
\end{equation}
YCbCr is widely used for skin color detection due to its ability to better represent color images with uneven illumination \cite{chai2000bayesian}. This helps in separating color from the illumination information. 

We found that representing the image in the YCbCr color system during the scoring process improves the correlation between human and machine ratings. The human eyes are more sensitive to black/white changes in the image rather than to color. This in turn affects the human judgment to be based more on geometrical changes of the face attributes rather than color changes. Thus, we transform the image from the RGB into the YCbCr color space to better capture the black/white information. 
\begin{figure}[h]
\begin{center}
\fbox{\includegraphics[width=13cm]{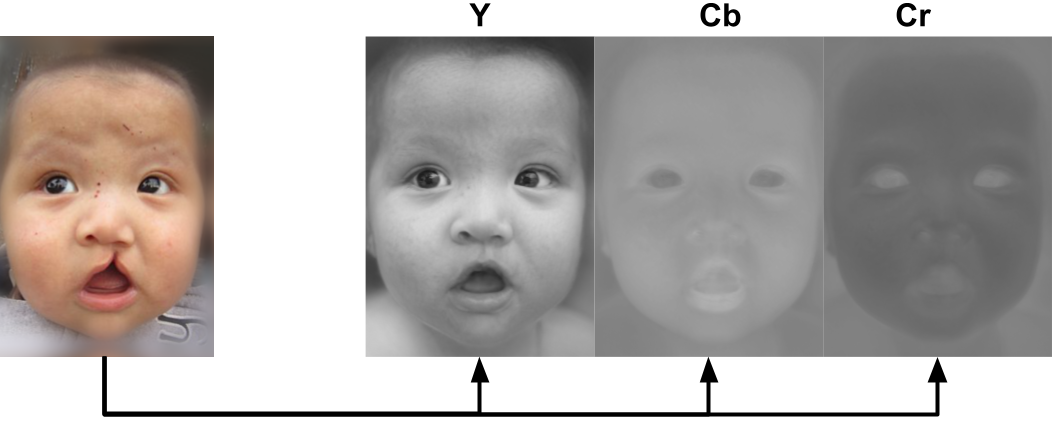} }
\end{center}
\caption{Transforming the face image from RGB to YCbCr. Note that most of the variation is in the Y component. Cb and Cr contain negligible variation being  only closer to a constant value. }
\end{figure}

\subsection{Heatmap Generation via Pixelwise Squared Error}

All the steps in the proposed method collaboratively contribute to the proposed scoring system. Heatmap generation plays a key role in defining the anomaly score. It is important for the heatmap to highlight the anomalous parts rather than the actual pixel difference. Generally, the heatmap can be created by comparing the original image with the normalized one by means of a similarity measure. Several similarity measures can be used to generate the heatmap. These measures can be divided into two main categories:  shallow and deep measures, respectively. The first category uses an explicit definition of the difference between the two images. One example is the simple Pixelwise Subtraction Error (PSE) which measures the difference in the color intensity between two images $x_{org}$ and $ x_{norm}$ as  follows:
\begin{equation}
\label{eqn:PSE}
    PSE(x_{org},x_{norm})=(x_{org}-x_{norm})\bigodot (x_{org}-x_{norm}) ,
\end{equation}
where $\bigodot$ denotes the Hadamard product \cite{horn1990hadamard}. In essence, the majority of  similarity indexes employ  some type of heatmap to calculate the index. Structural Similarity Index Measure (SSIM) \cite{wang2004image} is another popular similarity index that compares the difference of quality between  two images in terms of color and contrast as well as  structural differences. It also generates a heatmap depending on the above mentioned difference features. It is interesting to observe that  deep similarity measures have incorporated an   implicit definition of the difference between the two images. This can be advantageous for conducting the comparison because usually there is no proper model to capture the actual difference. Instead of employing an explicit definition, the deep comparison methods employ  features extracted by deep neural networks pretrained on a dataset of different types of objects, i.e.,  they behave like objects feature extractors. An example of a deep comparison method  is LPIPS, an approach introduced earlier and that presents the additional feature of providing a heatmap before producing the loss. In general, almost all the similarity measures rely on heatmaps to calculate the similarity score. This make it possible to test different methods in terms of their abilities to relate the heatmaps with consistent scoring systems. 

In our framework, we use the PSE similarity measure by considering the squared difference between each pixel in the original image and the corresponding pixel in the normalized version. The pixel-based distance measure includes both the real anomalous difference and the difference that is caused by the change in illumination. We can consider the small changes in illumination  as an artifact since the human eyes are less sensitive to these small values. We found that, despite its simplicity, the PSE heatmaps are more intuitive and consistent with the generated scores than other types of heatmaps if supported with additional image processing techniques. On the other hand, heatmaps generated by LPIPS and SSIM do not greatly improve even if supported by additional image processing techniques.

\subsection{Morphological Erosion for Noise Reduction}
The normalized version of the face is not an exact copy of the original one as it includes  lighting changes and differences in the small fine details of the texture in the two images. To get rid of these artifacts, we apply morphological erosion \cite{serra2012mathematical} on the image in order to reduce the effect of the noise on the overall anomaly score.

The essence of the erosion process is to reduce the object boundaries and expand the size of the holes. This is done by finding the minimum value of the neighborhood pixels at a specific location of the image. Let $F(j,k)$ represent  the pixel value of the grayscale image $F$ under process at location $(j, k)$. The erosion process over a 3 x 3 pixel neighborhood is implemented by means of this transformation:
\begin{equation}
G(j, k) = MIN\{F(j, k), F(j, k + 1), F(j- 1, k + 1), ..., F(j + 1, k +1)\} .
\end{equation}

This process is repeated around the image portion that is subject to  analysis. We apply the erosion process on all the Y, Cb and Cr image channels as we assume that the noise is contained in all the channels but with higher magnitude in the Y channel.

\subsection{Anomaly Score Calculation}
Given the post-processed abnormality heatmap from the previous step, we mask the unwanted parts of the image using a mask image $m$, sum up all the remaining pixels and divide by their count $N$. This number represents an overall anomaly score indicating how much energy is contained in the heatmap. The higher the number, the more sever the facial abnormality is. 
\begin{equation}
S=-log(\frac{||PSE(x,y)||_F}{N})=-log(\frac{\sqrt{\sum_{i,j}{(x_{i,j}- y_{i,j})^2}}}{N})  . 
\end{equation}

In the evaluation phase, the score calculation is conducted  for all other types of heatmaps under comparison (LPIPS, SSIM instead of PSE). 
LPIPS represents  another widely used option to measure the difference between  two images. Furthermore, it has the ability to semantically localize the difference between images. However, we show in the evaluation section that its performance is poor compared to the simple difference map. The simple average PSE loss yields  closer scores to the human survey compared to LPIPS and other well known losses. 

\section*{Results and Discussion}
\subsection{Method Evaluation Criteria}
To evaluate our proposed framework, thirteen raw images of individuals with cleft lips were used as evaluation inputs for our method. The latent vector most closely matching the original deformed face was then retrieved. An iterative algorithm was utilized to jointly update the latent vector and optimize the weights of the generative model so as to more closely represent distinctive features of the raw image. The iterative algorithm was carried out through back propagation of the difference loss between the abnormal face and the image generated in the preceding step. This  operation was repeated for a sufficient number of iterations (50) to grossly obtain the desired facial details absent in the original anomalous features (i.e., the “normal transformation”).

Pixelwise subtraction in the YCbCr color space between the native abnormal image and its transformed counterpart allows  calculation of the  anomaly score and  generation of the anomaly heatmap. For the cohort of cleft faces, the oral/nasal region of interest for each image was isolated for analysis by manually masking the remainder of the face. The relationship between human ratings of the 13 cleft images and our machine-genera- ted scores was assessed using  Pearson’s correlation.

\textbf{Collecting human ratings:} Scores generated by the proposed method were compared to human ratings obtained using a survey of 80  raters evaluating 13 cleft lip images depicting a range of deformity scores from 1 for most abnormal to 7 for almost normal.

\begin{figure}[h]
\begin{center}
\fbox{\includegraphics[width=9cm]{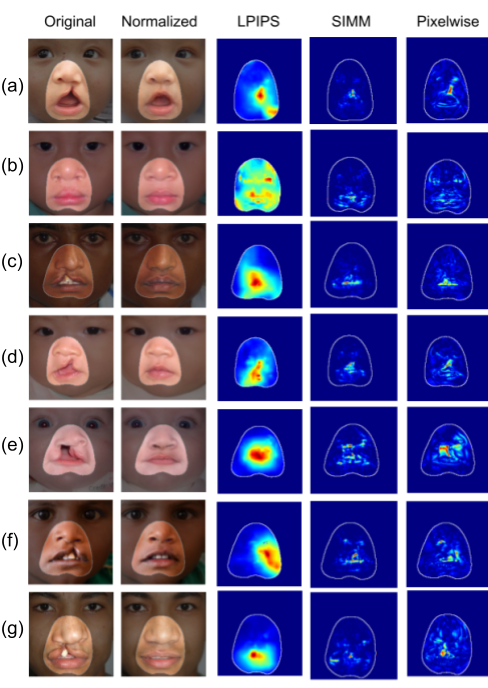} }
\end{center}
\caption{\label{img:heatmaps}Comparison of various  heatmaps generated by different similarity indexes. Facial images in rows (a-g) retrieved  from \cite{1png}, \cite{19png},\cite{29png},\cite{12png},\cite{2png},\cite{13png},\cite{4png}, respectively.}
\end{figure}

\textbf{Computing resources:} We  conducted the analysis on Python 3.6 using Pytorch, Opencv, Scipy and Skimage libraries on Intel i7-10751H CPU 2.6 GHz with Nvidia GeForce 2080 Super with Max-Q design.\\
\textbf{Data availability:} The 13 images that were used and measured in the study are all open source and licensed for re-use and the references for all 13 images have been included. So those 13 individuals did not require parental consent. The human raters were simply asked to give a 1-7 rating of the 13 open-source images and our institutions did not require an IRB approval for that. 
All remaining images are either wholly fabricated from StyleGAN2, or transformed versions of the same 13 open-source images.\\
Examples of heatmaps are presented in Figure \ref{img:heatmaps}. For each case, we present the original, the normalized and the abnormality heatmap. Everything other than mouth and nose parts are dimmed. The heatmap highlights the abnormal regions with brighter pixels meaning that the level of abnormality is higher than the regions with darker pixels. 
For the heatmaps generated by the PSE,  bright pixels covering the cleft region are illustrated in the majority of images. In addition, certain pixel locations are also highlighted but in lower amount. For the LPIPS case, the  identified regions in the heatmaps are generally wide and not localised. Also, some locations were selected as  anomalous although they are normal and even smooth. This leads to the conclusion that the deep features used in the LPIPS similarity index still suffer from limited anomaly detection quality. SSIM can generate more accurate anomaly heatmaps, but still does not  highlight the anomaly as a contiguous blob. This is because it focuses on the structural details of the images rather than the general large changes in color. An important feature of the PSE heatmap is the fact that it  was able to highlight the structural changes as well as the large contiguous blobs. Therefore, it turns out that the PSE heatmap is closer to the human intuition than the other generated heatmaps.\\

Regarding the correlation between the human survey and the generated scores, Table \ref{tab:correlation_table} shows 94\% correlation for the heatmaps generated by PSE for the mouth/nose parts and 91\% for scoring the entire face except the eyes. This correlation was achieved when the complete framework was  implemented, including the  combination of erosion and color transformation operations. This means that despite some anomaly heatmaps generated by the PSE are not perfect, the assumptions regarding the noise modeling in our proposed framework are correct in the sense that the additional post-processing operations were responsive and well integrated with the PSE heatmaps. Interestingly, these operations slightly improve  the correlation (74.1 to 76.9\%) in case of LPIPS. This is due to the fact that the highlighted anomaly regions were not accurate and noisy. 
Similarly, SSIM scores in Table \ref{tab:correlation_table} show the worst performance and correlation degradation (from 63 to 60\% in case of YCbCr transformation) and slight improvement in case of adding the erosion operation (63 to 65.6). In general, slight performance improvement can be gained by adding any combinations of these image processing operations.  
\begin{table}[!ht]
\begin{adjustwidth}{0in}{0in} 
\centering
\caption{\label{tab:correlation_table}Correlation coefficient between the human and machine scores of the 13 faces under analysis, utilizing different heatmap generation approaches.}
\begin{tabular}{ |c||c|c|c|c||c|c|c|c|  }

\hline
With Adaptation&
\multicolumn{4}{c||}{Mouth Only} &\multicolumn{4}{c|}{Entire Face}\\
\hline
Heatmap& Base & +YCbCr &+Erosion&+Both& Base & +YCbCr &+Erosion&+Both\\
\hline
LPIPS & 74.1&77.5&74.3&76.9&58.7&77.1&59.1&74.7 \\\hline
PSE & 72&83.1&79.7&94.2&55.9&78.7&74.5&91.1\\\hline
SSIM &63&60&65.6&55.2&57.7&47.3&72&45.7 \\
\hline

Without Adaptation&\multicolumn{4}{c||}{Mouth Only} &\multicolumn{4}{c|}{Entire Face}\\
\hline
LPIPS & 58.8&67.6&57.6&71.9&62.1&49.9&61&54.4 \\\hline
PSE & 56.8&77.5&60.3&79.9&43.6&54.6&0.44&51.4\\\hline
SSIM &48.1&48.1&42.3&47.2&57.9&40.4&52.3&40.6 \\
\hline

\end{tabular}
\label{table1}
\end{adjustwidth}
\end{table}

The reason for having better correlation of the mouth/nose parts compared to the entire face can be either due to the fact that the amount of noise introduced from different sources is proportional to the number of pixels considered in analysis or the proposed normalization phase did a better job of normalizing the mouth/nose than other face parts as well as the overall image appearance reconstruction. 
Regarding eyes normalization, it is very hard for the normalization phase to refine or reconstruct it. This may be explained by the details of the eyes which are more random than other face parts.  The eye pixels capture random reflections of the surrounding environment. These reflections are difficult to  capture by deep models including the StyleGAN2 generator. The bottom part of Table \ref{table1} shows the correlation between the machine predictions with the human scores, in the situation when the proposed StyleGAN2 model adaptation  is not applied during the normalization phase. Clearly, the correlation is worse as a lot of facial details in the original image where not represented in the normalized face. This explains the importance of applying extra adaptation steps for  the StyleGAN2 facial generator model.

\begin{figure}[h]
\begin{center}
\fbox{\includegraphics[width=12.25cm]{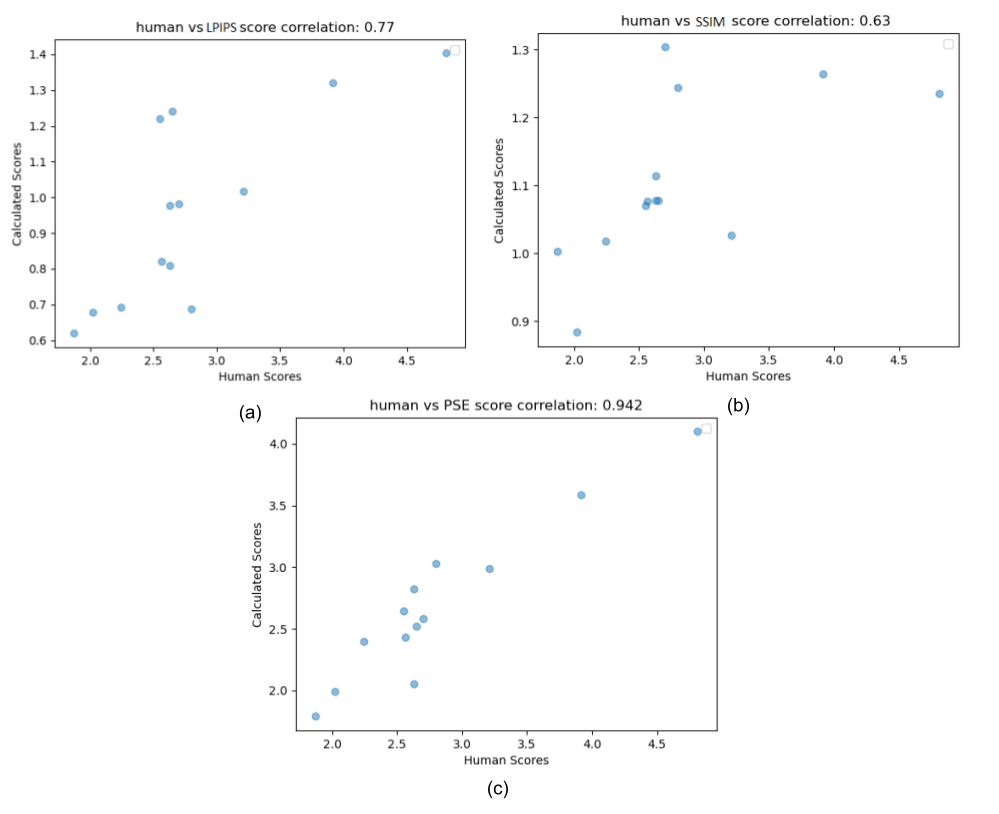} }
\end{center}
\caption{\label{img:scores}Correlation between human and machine scores from different algorithms.}
\end{figure}

Also, we evaluate in our framework the generated scores visually by comparing them in various scenarios with different similarity measures. Figure \ref{img:scores} shows a demonstration of the correlation between the human survey and various generated scores for different corresponding heatmaps. It is obvious that the PSE scores present better consistency than LPIPS and SSIM scores. The PSE scores are related linearly with the human scores. This was not possible before taking the log transform for the squared pixel differences during the scores calculation. This is important as the human judgment of the anomaly is very sensitive to small anomaly changes which makes the human penalize the small anomalies more than large ones. To cancel this non-linearity, we take the log of the scores.

The computer simulations illustrate that the method is capable to score each pixel of the original abnormal face if it represents a part of an abnormal region. Also, the Table \ref{tab:correlation_table} shows the superiority of the simple PSE among the LPIPS and SSIM measures.
The computation time required to evaluate the score of one face image amounts to 50 seconds. The proposed method assumes   2,000 face inversion iterations, 50 adaptation iterations and all the other previously mentioned processing steps. The proposed method  presents lower computational  overhead relative to the approach in  \cite{boyaci2020personalized},  which requires 9 minutes to score one image. 

\section*{Conclusions}

Localizing the abnormality regions in a patient’s face can be conducted  using the power of Generative Adversarial Networks (GANs). Even if there is no representation in the GAN pretrained model, we were able to embed  these details into the StyleGAN2 model using the classical backpropagation algorithm for some of the face details. This helped us to preserve the identity of the patient’s face for the generated image as well as to remove the abnormality regions and replace them with normal face attributes. Therefore, we were able to generate a heat map that highlighted the possible abnormal face parts using the pixel wise difference between the original abnormal face and the normalized one.








\section*{Acknowledgments}
This publication was made possible by NPRP13S-0127-200182 from the Qatar National Research Fund (a member of Qatar Foundation). The statements made herein are solely the responsibility of the authors.

%
%
%

\bibliographystyle{acm}
\bibliography{sample}

\end{document}